%% file: main.tex
\documentclass[runningheads]{llncs}

\usepackage{eccv}

\usepackage{eccvabbrv}

\usepackage{graphicx}
\usepackage{booktabs}
\usepackage{amsmath,amssymb,mathtools,bm}
\usepackage{subcaption}
\usepackage{pifont}
\usepackage{multirow}
\usepackage[table]{xcolor}
\usepackage{float}

\usepackage[accsupp]{axessibility}  

\definecolor{lightrow}{gray}{0.93}

\newcommand{\NA}{\textemdash}

\usepackage{hyperref}

\usepackage{orcidlink}

\begin{document}

\title{Self-Attention And Beyond the Infinite:\texorpdfstring{\\}{: } Towards Linear Transformers with Infinite Self-Attention}

\titlerunning{Infinite Self-Attention}

\author{
Giorgio Roffo\inst{1}$^{*}$\orcidlink{0000-0003-4170-914X} \and
Hazem Abdelkawy\inst{2} \and
Nilli Lavie\inst{3} \and
Luke Palmer\inst{4}$^{*}$
}

\authorrunning{G.~Roffo et al.}

\institute{
Equixly API Security, Italy\\
\email{giorgio.roffo@equixly.com}
\and
Toyota Motor Europe, Belgium\\
\email{hazem.abdelkawy@toyota-europe.com}
\and
MindVisionLabs, UK\\
Institute of Cognitive Neuroscience, University College London, UK\\
\email{n.lavie@ucl.ac.uk}
\and
GlimpseML, UK\\
\email{luke@glimpse.ml}
}

\maketitle
\begingroup
\renewcommand\thefootnote{\fnsymbol{footnote}}
\footnotetext[1]{This work was initiated and primarily carried out while working at MindVisionLabs.}
\endgroup
\input{sec/0_abstract}
\input{sec/1_intro}
\input{sec/2_related}
\input{sec/3_method}
\input{sec/4_experiments}
\input{sec/6_conclusion}

\section*{Acknowledgements}
We gratefully acknowledge the support of Toyota Motor Europe (TME) and Equixly API Security for this work.

\bibliographystyle{splncs04}
\bibliography{main}

\clearpage
\input{sec/X_suppl}

\end{document}

%% file: sec/0_abstract.tex
\begin{abstract}
The quadratic cost of softmax attention limits Transformer scalability in high-resolution vision.
We introduce \textit{Infinite Self-Attention} (InfSA), a spectral reformulation that treats each attention layer as a diffusion step on a content-adaptive token graph, accumulating multi-hop interactions through a discounted Neumann series over attention matrices. This formulation connects self-attention to classical graph centrality measures---Katz, PageRank, and eigenvector centrality---yielding interpretable and structurally grounded token weighting. We further show that this Neumann kernel coincides with the fundamental matrix of an absorbing Markov chain~\cite{Roffo9119168,kemeny1960finite}, linking each token's centrality score to its expected number of random-walk visits before absorption.
Building on this, we propose \textit{Linear-InfSA}, an $\mathcal{O}(N)$ variant that approximates the principal eigenvector of the implicit attention operator without forming the $N \times N$ matrix. It maintains an auxiliary state of fixed size $\mathcal{O}(d_h)$---where $d_h$ is the per-head dimension, independent of the sequence length $N$---is drop-in compatible with standard Vision Transformers, and supports stable forward and backward passes at $4096^2$ resolution and inference at $9216^2$ ($\sim$332k tokens).
Integrated into a 4-layer ViT with 53.5M parameters and 59\,GFLOPs at $224^2$, Linear-InfSA achieves 84.7\% top-1 on ImageNet-1K, a $+$3.2\,pp purely architectural gain over a standard 4-layer ViT baseline (81.5\%) trained with an identical recipe. On ImageNet-V2, all InfViT variants surpass every compared baseline (up to 79.8\% vs.\ 76.8\% for the best prior method), indicating robust generalization under distribution shift. Attention quality evaluations confirm semantically grounded maps: MoRF-AOC reaches 76.0\% and bounding-box PR-AUC 76.1\%, versus 42.6\% and 56.2\% respectively for softmax ViT. In scalability benchmarks on an A100 40\,GB GPU, Linear-InfViT delivers 231\,img/s at 0.87\,J/img---a $13\times$ improvement in both throughput and energy over a standard ViT of equal depth---and is the only tested model to complete $9216^2$ inference without running out of memory. The linear approximation faithfully recovers the dominant eigenvector of the full quadratic operator (cosine similarity 0.985).
Code available at:
\href{https://huggingface.co/groffo/infinite-self-attention}{huggingface.co/groffo/infinite-self-attention}

\keywords{Graph diffusion attention \and Absorbing Markov chains \and Linear-complexity Vision Transformers \and Eigenvector centrality \and Interpretable attention maps}
\end{abstract}

%% file: sec/1_intro.tex
\section{Introduction}
\label{sec:intro}

Transformer architectures underpin modern vision~\cite{dosovitskiy2021vit, liu2022swinv2, rao2022hornet} and language models~\cite{vaswani2017attention,brown2020language,touvron2021training}, yet their quadratic attention cost limits scalability in high-resolution and long-context settings~\cite{wang2020linformer, performer}. This has motivated numerous efficient attention mechanisms~\cite{linformer, xiong2021nystromformer, dao2023flashattention, beltagy2020longformer}.  
This computational bottleneck also carries environmental cost: data-centre consumption is projected to nearly double by 2030~\cite{IEA2025EnergyAI,DOE2024USReport}, and quadratic attention dominates Transformer energy budgets~\cite{Strubell2019EnergyNLP}.

\begin{figure}[tb]
  \centering
  \includegraphics[width=0.85\linewidth]{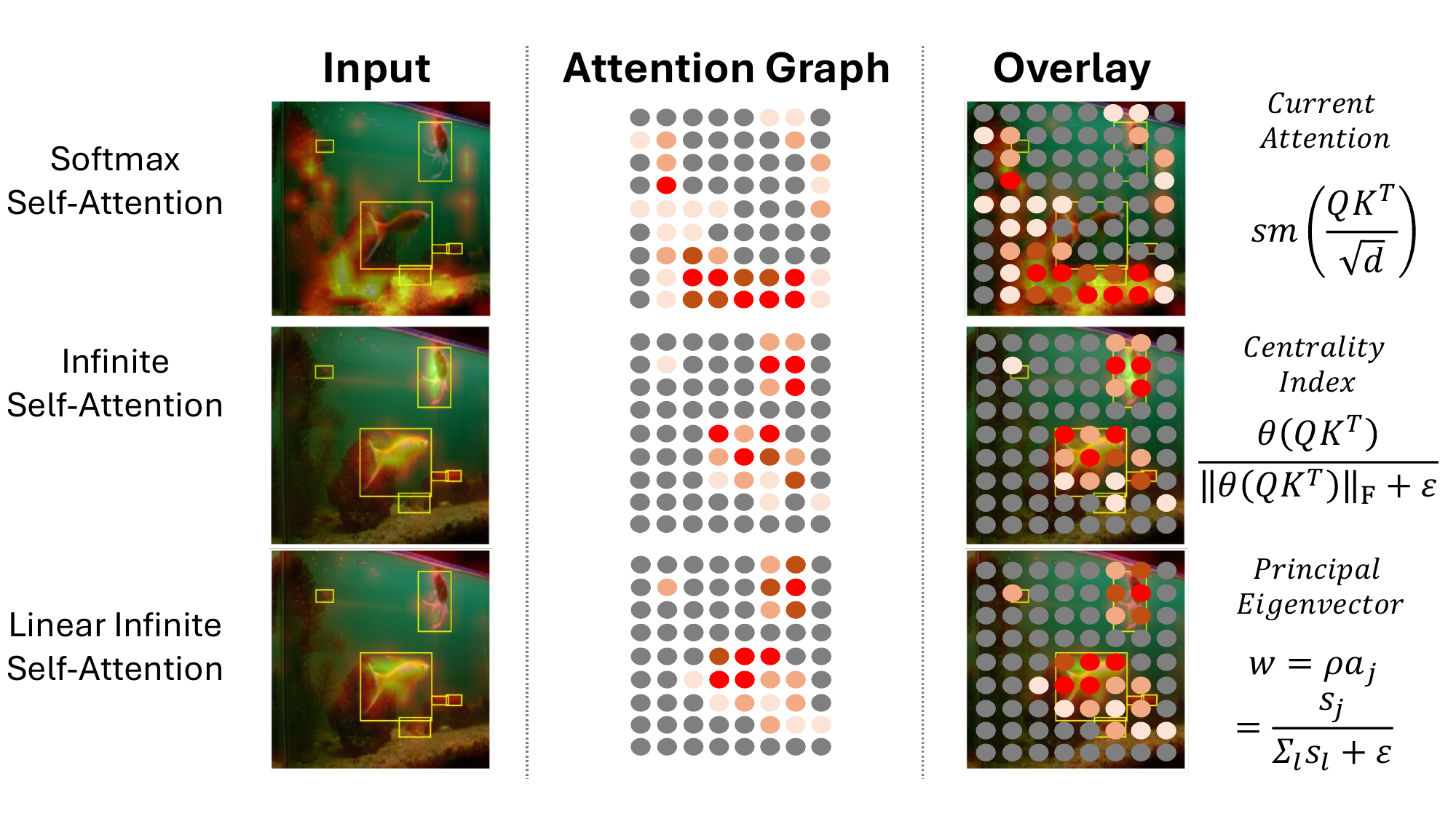}
  \caption{\textbf{Comparison of attention graphs.} Visualization of ViT-L/16 attention maps on ImageNet. Softmax attention distributes focus across background regions, while InfSA variants produce sharper, object-aligned activations.}
  \label{fig:comparison}
\end{figure}

Despite progress, most efficient variants approximate or sparsify the attention matrix without a principled model of token interaction. Standard Transformers aggregate dependencies implicitly across stacked layers, offering limited control over multi-hop influence or interpretability. Empirical analyses further show that attention weights may highlight diffuse or semantically irrelevant regions~\cite{abnar2020quantifying, chefer2021transformer}.  
Graph-theoretic formulations offer a more structured perspective: diffusion processes quantify node influence through centrality measures such as Katz~\cite{katz1953new}, PageRank~\cite{page1999pagerank, bianchini2005inside}, and eigenvector centrality~\cite{eigenvector_centrality}, enabling explicit multi-hop reasoning and structural interpretability---yet such principles remain underexplored within Transformer architectures.
Notably, the connection between infinite-path aggregation on weighted graphs and absorbing Markov chains was established by Roffo~\etal~\cite{Roffo9119168}, who showed that the fundamental matrix of a substochastic random walk yields the same Neumann kernel $N=(I-\gamma A)^{-1}$ used for centrality scoring, with entry $N_{ij}$ equal to the expected number of visits to node~$j$ before absorption when starting from node~$i$.

We introduce \textit{Infinite Self-Attention (InfSA)}, a spectral diffusion view of attention that aggregates information across layers through a truncated Neumann-series construction, akin to infinite-path kernels used in Katz/PageRank scoring~\cite{Roffo7410835,Roffo9119168}. The same Neumann kernel admits an absorbing Markov chain interpretation~\cite{Roffo9119168,kemeny1960finite}: tokens are transient states of a random walk on the attention graph, and their centrality scores correspond to expected visit counts before diffusion terminates. We also develop \textit{Linear-InfSA}, a scalable $\mathcal{O}(N)$ variant that approximates dominant attention directions via the $A^{k\to\infty}$ eigenvector limit (see Fig.~\ref{fig:comparison}). Integrated into Vision Transformers, InfSA delivers strong ImageNet-1K/V2 accuracy, sharper and more localized attention maps, and stable scaling to 4K--9K inputs; derivations appear in the supplementary material.

\noindent
\textbf{Contributions.}
\begin{itemize}
    \item We connect attention propagation to eigenvector dynamics and nonlinear Perron--Frobenius theory, offering a principled view of global token influence.
    \item We introduce \textit{InfSA}, a spectral generalization of self-attention via graph diffusion and Neumann-series path integrals, and show that the resulting attention graph admits an absorbing Markov chain interpretation in which token centrality equals expected random-walk visits before absorption.
    \item We propose \textit{Linear-InfSA}, an $\mathcal{O}(N)$ approximation that avoids attention matrix construction, with a fixed-size $\mathcal{O}(d)$ auxiliary state independent of $N$, enabling stable scaling to high resolutions.
    \item Experiments on vision tasks illustrate interpretability, robustness, and scalability; InfSA establishes a conceptual foundation for efficient AI architectures.
\end{itemize}



\noindent
The rest of the paper is organized as follows: Sec.~\ref{sec:Related} reviews related work on efficient attention, graph centrality, and state-space models; Sec.~\ref{sec:method} derives Pure InfSA from the path-integral perspective and its absorbing Markov chain interpretation, then introduces Linear-InfSA; Sec.~\ref{sec:experiments} evaluates scalability, attention quality, and classification performance; Sec.~\ref{sec:conclusion} concludes.

%% file: sec/2_related.tex
\section{Related Work}
\label{sec:Related}

\textbf{Efficient Attention.}
The quadratic cost of softmax attention has driven many sub-quadratic alternatives.
Linformer~\cite{linformer} projects keys to fixed rank; Performer~\cite{performer} uses kernel random features; SOFT~\cite{lu2021soft} applies softmax-free kernels; FLatten~\cite{han2023flatten} restores rank in focused linear attention; Agent Attention~\cite{han2024agent} bridges softmax and linear attention via agent tokens; MLLA~\cite{han2024mlla} recasts Mamba-style gating as linear attention; and Fastformer~\cite{wu2021fastformer} replaces pairwise attention with additive pooling. FlashAttention~\cite{dao2023flashattention,dao2023flashattention2} optimizes softmax via fused kernels but retains $\mathcal{O}(N^2)$ compute. All approximate or sparsify the attention matrix without modeling multi-hop token influence. Our Linear-InfSA computes token centrality in $\mathcal{O}(N)$ with an $\mathcal{O}(d)$ state independent of $N$, while remaining drop-in compatible with ViT blocks.

\noindent\textbf{Graph Centrality and Markov Interpretations.}
Attention has been linked to graph reasoning~\cite{wang2018non,romero2021geometric}, but few works model explicit token diffusion. We formalize attention as a content-adaptive affinity graph connecting it to Katz centrality~\cite{katz1953new}, PageRank~\cite{page1999pagerank,bianchini2005inside}, and Infinite Feature Selection (Inf-FS)~\cite{Roffo7410835,Roffo9119168}, whose Neumann kernel coincides with the fundamental matrix of an absorbing Markov chain~\cite{kemeny1960finite}. TokenRank~\cite{erel2025attention} independently interprets attention as a Markov chain, computing the stationary distribution of a closed chain; InfSA instead computes the fundamental matrix $N{=}(I{-}\gamma\hat{A})^{-1}$ of an absorbing chain---encoding expected visit counts rather than steady-state probabilities. On interpretability, attention rollout~\cite{abnar2020quantifying}, attention flow~\cite{chefer2021transformer}, and perturbation saliency~\cite{fong2017interpretable} propagate relevance heuristically, while MoRF/LeRF~\cite{samek2016evaluating} evaluates faithfulness via masking. InfSA embeds centrality directly in the mechanism, unifying attention, diffusion, and attribution.

\noindent\textbf{State-Space and Convolution Models.}
SSMs such as Mamba~\cite{gu2023mamba} and RWKV~\cite{peng2023rwkv}, along with vision adaptations MambaVision~\cite{hatamizadeh2025mambavision} and HyenaPixel~\cite{spravil2024hyenapixel}, achieve sub-quadratic modeling via recurrence or long convolutions but do not derive token importance from graph centrality; InfSA's spectral formulation is complementary.

%% file: sec/3_method.tex
\section{Infinite Self-Attention (InfSA)}
\label{sec:method}

\textit{Infinite Self-Attention} (InfSA) generalizes Transformer self-attention~\cite{dosovitskiy2021vit} by modeling \textit{multi-hop dependencies} through a spectral, path-based interpretation of token interactions, inspired by Infinite Feature Selection (Inf-FS)~\cite{Roffo7410835,Roffo9119168}. This connects attention to classical \textit{graph diffusion} and \textit{ranking}---Katz, PageRank~\cite{katz1953new,page1999pagerank}---and to recent spectral views of attention~\cite{romero2021geometric,teo2024unveiling,kernelpca_attention,Roffo9119168}.
We derive a layer-wise formulation compatible with standard Transformer blocks and introduce \textit{Linear-InfSA}, a scalable $\mathcal{O}(N)$ variant that bypasses the $N{\times}N$ attention matrix while preserving the ViT residual structure~\cite{dosovitskiy2021vit}.

\subsection{Attention Graphs}
\label{sec:method:attgraph}

We interpret self-attention~\cite{vaswani2017attention,dosovitskiy2021vit} as a diffusion process on a fully connected, content-adaptive graph $\mathcal{G} = (\mathcal{V}, \mathcal{E})$ where each token is a node and each edge $(i,j)$ has weight $A_{ij} \geq 0$ derived from attention scores. Let $A \in \mathbb{R}^{N \times N}$ denote the attention matrix and $V \in \mathbb{R}^{N \times d}$ the value matrix. The self-attention update is a diffusion step:
\begin{equation}
    Y = AV,
    \label{eq:diffusion}
\end{equation}
where each output token aggregates values from all others, weighted by attention~\cite{buades2005non,wang2018non}. When $A$ is row-stochastic ($A\mathbf{1} = \mathbf{1}$), this is the random-walk operator on $\mathcal{G}$~\cite{ortega2018graph}.
Viewing $A$ as an affinity matrix connects attention to centrality measures---PageRank~\cite{bianchini2005inside}, Katz~\cite{katz1953new}, eigenvector centrality---and to Inf-FS~\cite{Roffo7410835,Roffo9119168}, which ranks features by structural importance on weighted graphs (see Fig.~\ref{fig:comparison}).

\subsection{Infinite Self-Attention (InfSA): Path Integrals on the Attention Graph}
\label{sec:infsa_full}

We extend the attention graph perspective by proposing \textit{Infinite Self-Attention} (InfSA)\footnote{The term ``Infinite'' refers to the limiting Neumann series, \ie, $\lim_{L\to\infty}\sum_{t=0}^{L}\gamma^{t}A^{t}$ and to the eigenvector limit $v=\lim_{k\to\infty}A^{k}x_{0}/\|A^{k}x_{0}\|_{1}$. Note: the actual computation is finite. We use $\gamma$ for the decay (discount) factor and reserve $\rho(\cdot)$ exclusively for the spectral radius of a matrix.}, a formulation inspired by the infinite path integration approach introduced by Roffo~\etal~\cite{Roffo7410835,Roffo9119168}. That approach ranks features by aggregating the weights of \emph{all} paths on a feature-affinity graph, summing matrix powers $A + A^2 + \cdots$ and closing the series via $(I{-}rA)^{-1}{-}I$. We map this construction onto self-attention: tokens replace features, the attention matrix replaces the feature-affinity matrix, and the resulting infinite-path scores become token centralities.

\noindent\textbf{Paths on the token graph.}
Let $\pi = (v_0{=}i,\, v_1,\, \dots,\, v_{t-1},\, v_t{=}j)$ denote a path of length $t$ from token $i$ to token $j$ in the attention graph $\mathcal{G}$. The \emph{path weight} is the product of attention weights along edges:
\begin{equation}\label{eq:path_weight}
w(\pi) \;=\; \prod_{k=0}^{t-1} A(v_k, v_{k+1}).
\end{equation}
This product measures the cumulative affinity along the path: it is high when all consecutive tokens are mutually relevant.

\noindent\textbf{Aggregation over all paths of fixed length.}
Let $\mathbb{P}_{ij}^{t}$ be the set of all paths of length $t$ from $i$ to $j$. The total contribution of length-$t$ paths is
\begin{equation}\label{eq:path_sum}
R_t(i,j) \;=\; \sum_{\pi \in \mathbb{P}_{ij}^{t}} w(\pi).
\end{equation}
By standard matrix algebra~\cite{horn2012matrix}, this equals the $(i,j)$-entry of the $t$-th power of $A$:
\begin{equation}\label{eq:path_power}
R_t \;=\; A^{\,t}.
\end{equation}
In other words, $A^t(i,j)$ aggregates the contributions of \emph{all} length-$t$ walks from token $i$ to token $j$ on the attention graph.

\noindent\textbf{Token score at a fixed path length.}
To measure the importance of token $i$ at path length $t$, we marginalize over all destination tokens:
\begin{equation}\label{eq:token_score_t}
c_t(i) \;=\; \sum_{j=1}^{N} A^t(i,j).
\end{equation}
The score $c_t(i)$ quantifies how much token $i$ participates in \emph{all} subsets of $t$ interacting tokens: the higher the score, the more central token $i$ is at this interaction depth.

\noindent\textbf{Aggregation over all path lengths.}
A complete assessment of token importance accounts for interactions at all depths $t=1,2,\dots$\ However, the unregularized sum $\sum_{t\geq 1}A^t$ may diverge when $\rho(A)\geq 1$. Following~\cite{Roffo7410835,Roffo9119168}, we introduce a discount factor $\gamma\in(0,\,1/\rho(A))$ that geometrically attenuates longer paths:
\begin{equation}\label{eq:regularized_score}
\check{c}(i) \;=\; \sum_{t=1}^{\infty} \gamma^{\,t}\, c_t(i) \;=\; \sum_{t=1}^{\infty} \sum_{j=1}^{N} \gamma^{\,t}\, A^t(i,j).
\end{equation}

\noindent\textbf{Closed-form via geometric matrix series.}
The regularized infinite-path matrix is $\check{C} = \sum_{t=1}^{\infty} (\gamma A)^t$. By the convergence property of geometric power series of matrices~\cite{horn2012matrix}, if $\gamma < 1/\rho(A)$ then $\rho(\gamma A) < 1$ and:
\begin{equation}\label{eq:closed_form}
\check{C} \;=\; (I - \gamma A)^{-1} - I.
\end{equation}
The proof relies on Gelfand's formula: $\rho(\gamma A) = \rho((\gamma I)\, A) \leq \rho(\gamma I)\,\rho(A) = \gamma\,\rho(A) < 1$, which guarantees $\lim_{t\to\infty}(\gamma A)^t = 0$ and hence absolute convergence of the series~\cite{horn2012matrix,Graham:1994}. The matrix $\check{C}$ encodes the cumulative, discounted influence of every token on every other across all interaction depths.

When $\rho(\gamma A)<1$, the discounted path sum $\check{C}$ has a natural probabilistic reading: it coincides with the fundamental matrix of an absorbing Markov chain~\cite{Roffo9119168,kemeny1960finite}. In this view, tokens are transient states of a random walk on the attention graph, with a complementary absorption probability at each step; entry $\check{C}_{ij}+\delta_{ij}$ equals the expected number of visits to token~$j$ before absorption starting from~$i$. We develop this interpretation fully in Sec.~\ref{sec:markov}.

\noindent\textbf{Token centrality scores.}
The final per-token score is obtained by marginalizing over destinations:
\begin{equation}\label{eq:final_score}
\check{c}(i) \;=\; [\check{C}\,\mathbf{e}]_i \;=\; \bigl[(I - \gamma A)^{-1}\,\mathbf{e}\bigr]_i - 1.
\end{equation}
Ranking tokens in decreasing order of $\check{c}$ yields a principled ordering by structural importance in the attention graph, where the most influential tokens---those that participate in many high-weight, multi-hop interactions---appear at the top.

\noindent\textbf{Layer-wise implementation.}
In a Transformer with $L$ layers, $A^{(l)}$ denotes the attention matrix at layer $l$. InfSA accumulates post-attention outputs with geometric decay:
\begin{equation} \label{eq:infsa_series_layerwise} 
S_L \;=\; \sum_{t=1}^{L} \gamma^{\,t}\big(A^{(t)}\!\cdots A^{(1)}\big)X^{(0)}, 
\end{equation}
where each layer adds progressively longer effective paths. Standard self-attention is the $L{=}1$ case.
When $A^{(l)}{=}A$ for all $l$, the partial sum approximates the Neumann-series identity:
\begin{equation}
\label{eq:neumann}
\sum_{t=1}^{\infty} \gamma^{\,t}A^{t} \;=\; (I - \gamma A)^{-1} - I, \qquad \gamma < 1/\rho(A),
\end{equation}
which coincides with Eq.~\ref{eq:closed_form}.
In the heterogeneous case, Frobenius normalization ($\|\hat{A}^{(l)}\|_F{=}1$, Eq.~\ref{eq:frob_norm}) ensures $\|S_L\|\leq\gamma/(1{-}\gamma)$ for $\gamma<1$, guaranteeing bounded outputs regardless of per-layer variation.
Since the series is truncated at depth $L$, $\gamma$ also serves as a tunable design choice modulating the contribution of deeper layers.

\noindent\textbf{Matrix properties and normalization.}  
To construct a positive operator at each layer, we use $\tilde A = \phi(QK^\top)$ where $\phi = \text{ReLU}$ ensures non-negativity. This is followed by Frobenius normalization:
\begin{equation}
\label{eq:frob_norm}
\hat A = \frac{\tilde A}{\|\tilde A\|_F + \varepsilon},
\end{equation}
where $\|\cdot\|_F$ is the Frobenius norm, and $\varepsilon$ prevents division by zero. This yields $\hat A \geq 0$ with bounded energy.

\noindent\textbf{Per-layer output.}
At each layer~$l$, Pure InfSA computes:
\begin{equation}
\label{eq:infsa_layer_output}
Z^{(l)} = \hat{A}^{(l)} V^{(l)}, \qquad \hat{A}^{(l)} = \frac{[Q^{(l)}{K^{(l)}}^\top]_+}{\|[Q^{(l)}{K^{(l)}}^\top]_+\|_F + \varepsilon},
\end{equation}
replacing the standard softmax$(QK^\top/\sqrt{d_k})V$ in conventional attention.
The accumulated output across all $L$ layers is $S_L = \sum_{l=1}^{L}\gamma^{\,l}Z^{(l)}$.

Unlike softmax, which yields row-stochastic matrices ($A\mathbf{1}{=}\mathbf{1}$) and causes oversmoothing by mixing toward a stationary distribution~\cite{li2018deeper,oono2020graph}, Frobenius normalization bounds the total matrix energy $\|\hat A\|_F{=}1$, providing a sufficient condition for the operator to be contractive ($\rho(\hat A){<}1$). This ensures the discounted series $\sum\gamma^t\hat A^t$ remains convergent and transforms each layer from a probability-mixing step into a \textbf{centrality computation}: tokens are weighted by their structural importance---akin to Katz centrality~\cite{katz1953new}---rather than by local probability mass. The absorbing Markov chain interpretation (Sec.~\ref{sec:markov}) relies directly on this sub-stochasticity.

In this formulation, $t$ represents the hop count in the attention graph---equivalently, the length of token-to-token paths; $\gamma$ defines the horizon or decay of long-range effects; and $L$ is the Transformer depth, acting as a truncation point for the infinite sum. Each layer $l$ thus approximates the $l$-th power $A^l$ of the underlying attention operator, and the accumulation (Eq.~\ref{eq:infsa_series_layerwise}) implements an explicit integration over all paths of length up to $L$, embedding graph diffusion directly into Transformer computation.

\begin{figure}[tb]
  \centering
  \begin{subfigure}[t]{0.4\linewidth}
    \centering
    \includegraphics[width=\linewidth]{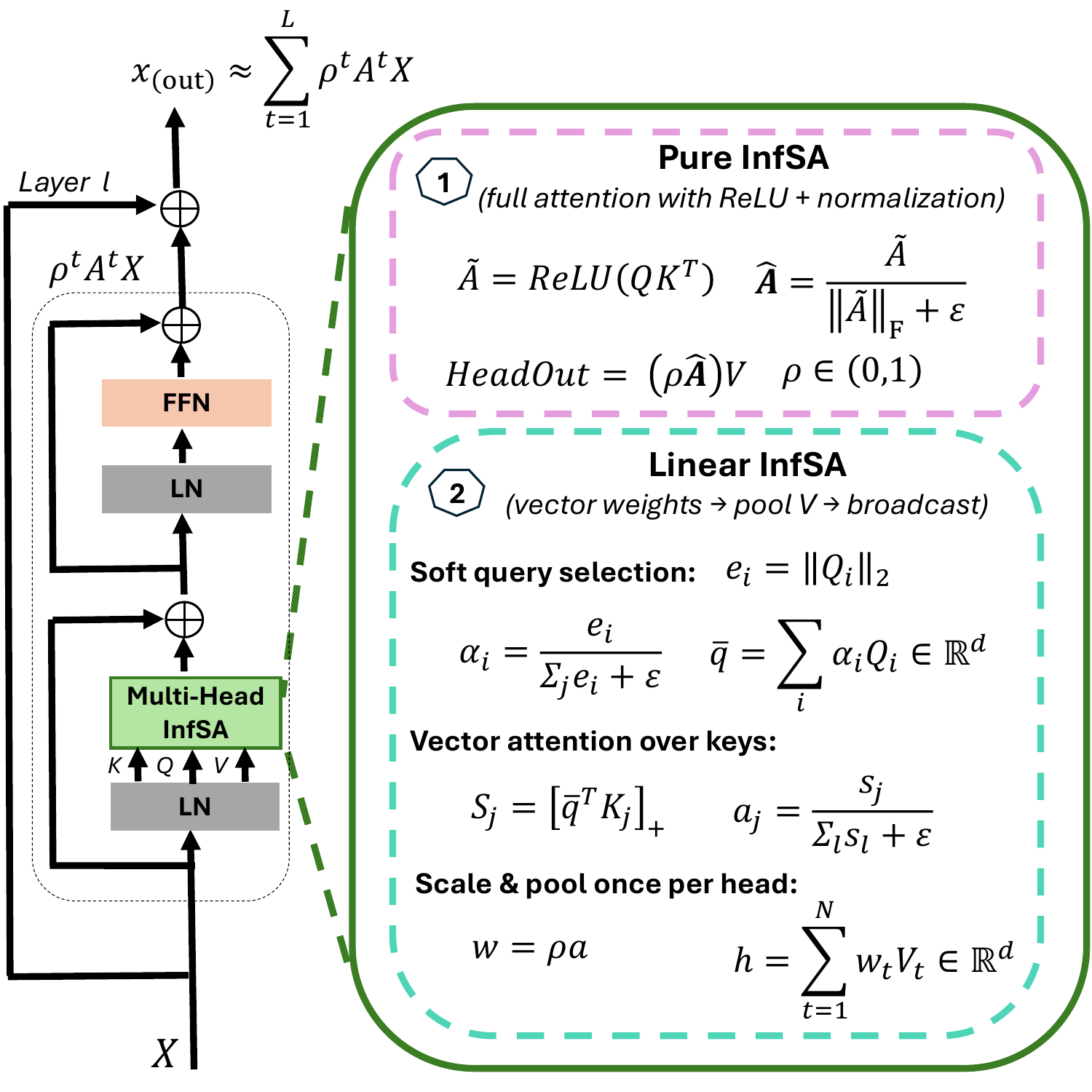}
    \caption{}
    \label{fig:method}
  \end{subfigure}\hfill
  \begin{subfigure}[t]{0.47\linewidth}
    \centering
    \includegraphics[width=\linewidth]{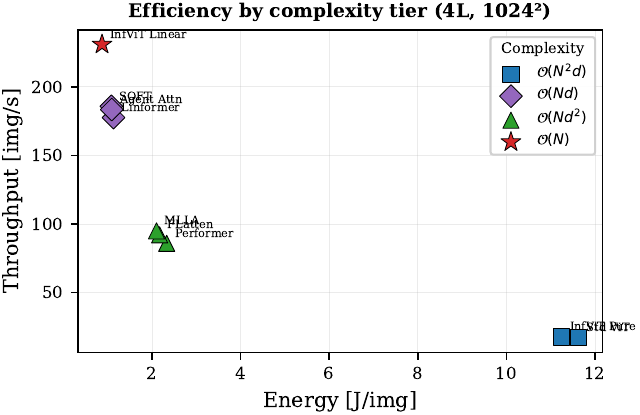}
    \caption{}
    \label{fig:complexity-tiers}
  \end{subfigure}
  \caption{\textbf{(a) InfSA in a Pre-LN ViT block.}
    Two InfSA variants within standard Transformer scaffolding:
    (1)~\textit{Pure InfSA} uses full attention with ReLU and Frobenius normalization,
    accumulating discounted outputs across layers;
    (2)~\textit{Linear InfSA} computes soft token scores, pools values per head, and broadcasts
    context with per-layer scaling. Both are drop-in compatible with Transformer blocks.
    \textbf{(b) Efficiency by complexity tier (4L, $\mathbf{1024^2}$).}
    Inference throughput vs.\ energy per image for nine attention mechanisms, colored by asymptotic complexity.
    InfViT Linear ($\mathcal{O}(N)$, red star) achieves the highest throughput at the lowest energy cost.}
  \label{fig:method-and-complexity}
\end{figure}

The exponential decay $\gamma^l$ introduces an explicit depth bias that limits over-propagation and helps avoid oversmoothing~\cite{li2018deeper,oono2020graph}: earlier layers contribute more strongly, while deeper paths are attenuated. If $\gamma$ is \emph{learned} (constrained to $(0,1)$ via sigmoid), it adaptively tunes the integration horizon per head. We implement InfSA within a Pre-LN Transformer (see Fig.~\ref{fig:method}$\langle$1$\rangle$) by computing $\hat A$ at each layer and accumulating outputs $S_L=\sum_{l=1}^{L}\gamma^{\,l}Z^{(l)}$.

\subsection{Absorbing Markov Chain Interpretation of InfSA}
\label{sec:markov}

The Neumann-series formulation of Pure InfSA (Eq.~\ref{eq:neumann}) admits a direct probabilistic interpretation through absorbing Markov chains~\cite{Roffo9119168,kemeny1960finite,seneta2006nonnegative}. This connection shows that InfSA computes expected visit counts in a random walk over the token graph, linking self-attention to classical stochastic processes.

\begin{figure}[tb]
  \centering
  \includegraphics[width=0.8\linewidth]{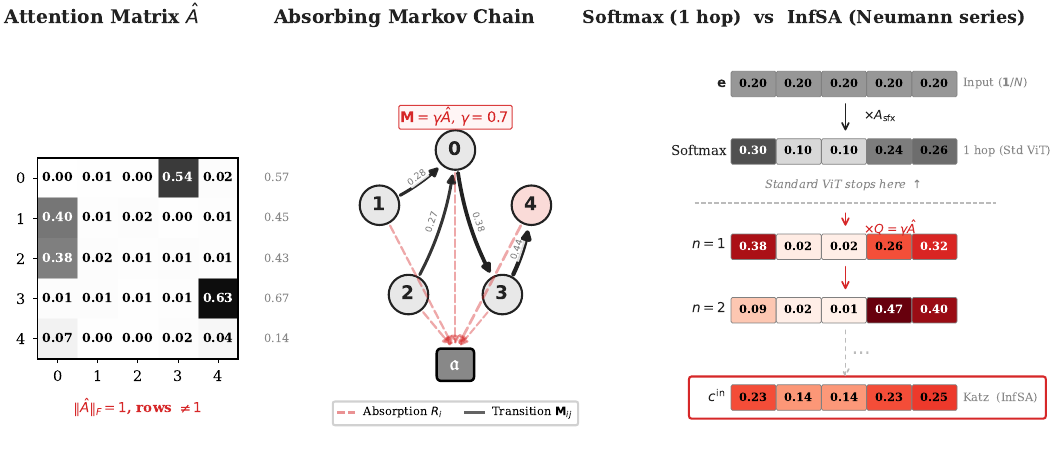}
  \caption{\textbf{Softmax attention (1-hop) vs.\ InfSA (Neumann series).}
  \textbf{Left:} Frobenius-normalized $\hat{A}$ ($\|\hat{A}\|_F{=}1$); row sums vary, unlike softmax.
  \textbf{Middle:} Absorbing Markov chain $\mathbf{M}{=}\gamma\hat{A}$; dashed red arrows show absorption $R_i{=}1{-}\sum_j \mathbf{M}_{ij}$ into~$\mathfrak{a}$.
  \textbf{Right:} Starting from the all-ones input $\mathbf{e}$ (Eq.~\ref{eq:final_score}):
  softmax attention (row-stochastic, 1-hop) ranks token~0 first via column sums, since many tokens directly attend to it.
  InfSA iterates $\mathbf{M}$ further; at $n{=}2$ the chain $0{\to}3{\to}4$ redirects mass to token~4, and the Katz centrality $c^{\mathrm{in}}$ correctly identifies token~4 as globally most important---the multi-hop outcome standard self-attention misses.}
  \label{fig:markov-higher-order}
\end{figure}

\noindent\textbf{Construction.}
Let $\hat{A} \in \mathbb{R}^{N \times N}_{\geq 0}$ be the Frobenius-normalized attention matrix (Eq.~\ref{eq:frob_norm}) and $\gamma \in (0, 1/\rho(\hat{A}))$ the decay factor.
Fig.~\ref{fig:markov-higher-order} illustrates the key intuition: a single power-method step (first-order attention) misidentifies the most important token, whereas iterating to the steady state---equivalent to the Neumann-series limit that InfSA computes---correctly reveals global structural importance through multi-hop propagation.
Define
\begin{equation}
\mathbf{M} = \gamma\, \hat{A}, \qquad
R_i = 1 - \gamma \sum_{j=1}^{N} \hat{A}_{ij},
\label{eq:markov_Q}
\end{equation}
where $\mathbf{M}_{ij} \geq 0$ represents the transition probability from token~$i$ to token~$j$, and $R_i \geq 0$ is the per-step absorption probability at token~$i$.
The matrix $\mathbf{M}$ is substochastic ($\mathbf{M}\mathbf{1} \leq \mathbf{1}$, with strict inequality for at least one row) whenever $\gamma \max_i \sigma_i < 1$, where $\sigma_i = \sum_j \hat{A}_{ij}$.
Frobenius normalization constrains $\|\hat{A}\|_F = 1$, which ensures $\rho(\mathbf{M}) = \gamma\,\rho(\hat{A}) < 1$.
We augment the $N$ token states with a single absorbing state~$\mathfrak{a}$, yielding the canonical form~\cite{kemeny1960finite}:
\begin{equation}
P = \begin{pmatrix} \mathbf{M} & R \\ \mathbf{0}^\top & 1 \end{pmatrix}
\in \mathbb{R}^{(N+1) \times (N+1)},
\label{eq:canonical_form}
\end{equation}
where $R = (R_1, \dots, R_N)^\top$ and $P$ is row-stochastic.
Tokens are transient states; the absorbing state $\mathfrak{a}$ represents termination of the diffusion process.

The fundamental matrix of this chain is
\begin{equation}
N = (I - \mathbf{M})^{-1} = (I - \gamma\, \hat{A})^{-1}
= \sum_{t=0}^{\infty} (\gamma\, \hat{A})^{\,t},
\label{eq:fundamental}
\end{equation}
which is precisely the Neumann kernel already underlying InfSA (Eq.~\ref{eq:neumann}).
The entry $N_{ij}$ has a concrete probabilistic meaning~\cite{kemeny1960finite,seneta2006nonnegative}: it equals the expected number of times the random walk visits token~$j$ before absorption, given that it starts at token~$i$.
The InfSA path-integral matrix $S = N - I = (I - \gamma\,\hat{A})^{-1} - I$ therefore counts expected visits \emph{excluding} the starting state.

\noindent\textbf{Token centrality as expected walk persistence.}
Row sums and column sums of $N$ yield two complementary centrality measures:
\begin{itemize}
\item \textbf{Outgoing influence:} $c_i^{\text{out}} = \sum_j N_{ij}$ is the expected total number of token visits before absorption when starting from token~$i$. Tokens with high outgoing influence initiate long, information-rich walks.
\item \textbf{Incoming centrality:} $c_j^{\text{in}} = \sum_i N_{ij}$ is the total expected visits to token~$j$ across all starting points. Tokens with high incoming centrality are structurally important in the attention graph.
\end{itemize}
These scores coincide with Katz centrality~\cite{katz1953new}, confirming that InfSA ranks tokens by global structural role rather than by local query--key affinity.

\noindent\textbf{Why Frobenius normalization enables absorption.}
Softmax normalization gives row-stochastic matrices ($\hat{A}\mathbf{1} = \mathbf{1}$), corresponding to a closed Markov chain with no absorbing state---the source of oversmoothing~\cite{li2018deeper}.
Frobenius normalization breaks row-stochasticity ($\rho(\hat{A}) < 1$ in practice), introducing a positive absorption probability at every step and ensuring convergence.
This absorbing-chain view is structurally identical to the Markov interpretation of Roffo~\etal~\cite{Roffo7410835,Roffo9119168}, where features are transient states and the fundamental matrix ranks them by expected walk persistence.  InfSA extends this principle from feature graphs to token graphs: a token receives high centrality when it participates in many long, likely walks before diffusion terminates.

\subsection{Linear-InfSA: Efficient Centrality Approximation}
\label{sec:linfsa}

To reduce the cost of explicit multi-hop accumulation, \textit{Linear-InfSA} approximates token centrality using the principal eigenvector of the implicit attention operator, without forming the full matrix $A$. Concretely, it approximates the dominant eigenvector of the Neumann kernel $\check{C} = (I - \gamma\hat{A})^{-1} - I$ computed by Pure InfSA (Eq.~\ref{eq:closed_form}), replacing the $\mathcal{O}(N^2)$ matrix inversion with a single $\mathcal{O}(N)$ power-iteration step. This yields a linear-time approximation of global influence, computed via simple vector operations and normalized iterations.
Let $X = [x_1, \dots, x_N]$ be the input tokens, and let $W_Q, W_V$ be the query and value projections. We tie projections and define $Q := X W_Q$, so that $Q = K$.
Tying $Q{=}K$ restricts the attention kernel to a symmetric similarity measure $(x_i^\top W^\top W x_j)$, forgoing asymmetric query--key interactions. This design choice is motivated by the eigenvector interpretation: the Perron eigenvector is a property of the symmetric operator $A + A^\top$, so symmetry is structurally appropriate. Asymmetric token interactions are recovered through the multi-head ensemble and the subsequent feed-forward network.

\noindent\textbf{Soft query construction.} We compute token-wise energies as the $\ell_2$ norms of the query vectors:
\begin{equation}
e_i = \|Q_i\|_2,
\end{equation}
providing a positive signal that reflects token prominence in embedding space~\cite{romero2021geometric}. These energies are normalized without softmax:
\begin{equation}
\label{eq:att_weights}
\alpha_i = \frac{e_i}{\sum_j e_j + \varepsilon}, \quad \text{with } \alpha \in \mathbb{R}^N_+, \quad \|\alpha\|_1 = 1,
\end{equation}
yielding a soft importance score that serves as a proxy for the dominant eigenvector of a positive operator~\cite{katz1953new,horn2012matrix}. No softmax or Frobenius normalization is required; the $\ell_1$ constraint ensures numerical stability.

\begin{figure}[tb]
  \centering
  \includegraphics[width=0.85\linewidth]{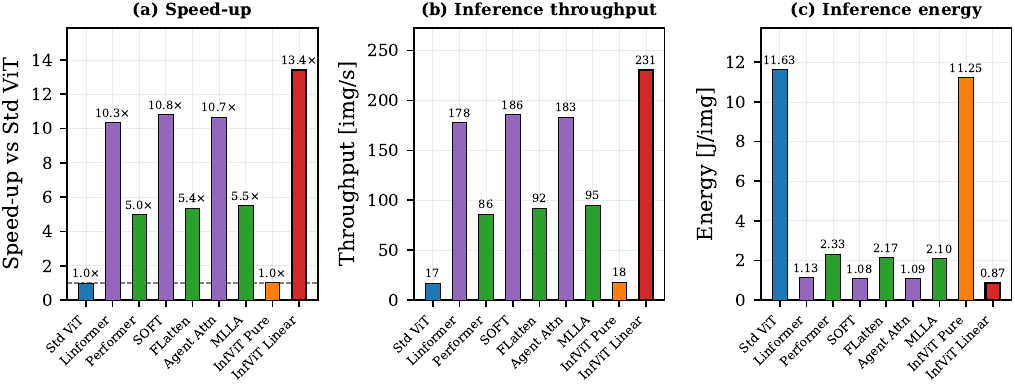}
  \caption{\textbf{Efficiency dashboard (4L-64H, inference at $\mathbf{1024^2}$).}
  Speed-up over Standard ViT, absolute throughput, and energy per image.
  InfViT Linear ($\mathcal{O}(N)$) achieves $13.4{\times}$ speed-up at 0.87\,J/img.}
  \label{fig:efficiency-dashboard}
\end{figure}

\noindent\textbf{Central query and attention over keys.} The soft central query is obtained via weighted averaging:
\begin{equation}
\bar{q} = \sum_i \alpha_i Q_i \in \mathbb{R}^d.
\label{eq:linfsa_qbar}
\end{equation}
We compute the scores over keys using a positive kernel:
\begin{equation}
S_j = [\bar{q}^\top K_j]_+,
\label{eq:linfsa_scores}
\end{equation}
and normalize again using an L1 constraint ($Q=K$):
\begin{equation}
a_j = \frac{S_j}{\sum_l S_l + \varepsilon}.
\label{eq:linfsa_iter}
\end{equation}
The Linear-InfSA weights implement a first-order Perron--Frobenius--style approximation of the dominant eigenvector of the implicit attention diffusion operator. Under nonnegativity (or strong positivity after a tiny floor), the iteration is an order-preserving, 1-homogeneous map whose normalized iterates converge to a unique eigenvector; in the strictly nonnegative case, it reduces exactly to the classical power method.
The resulting $\mathcal{O}(N)$ complexity yields a $13.4{\times}$ speed-up over Standard ViT, absolute throughput of 231\,img/s, and energy cost of only 0.87\,J/img at $1024^2$ (see the efficiency dashboard in Fig.~\ref{fig:efficiency-dashboard}).

\noindent\textbf{Final context pooling.} The head output is computed as a weighted sum of values:
\begin{equation}
h = \sum_{t=1}^N w_t V_t, \quad \text{with } w = \gamma a,
\end{equation}
mirroring the same scaling mechanism used in Pure InfSA. The resulting context vector $h$ is then \emph{broadcast} to all token positions and merged across heads (see Fig.~\ref{fig:method}$\langle$2$\rangle$).

\noindent\textbf{Query-independent weighting by design.}
The weight vector $a$ depends only on Keys (equivalently Queries, since $Q{=}K$) and is shared across all positions---a direct consequence of the eigenvector limit, where the operator is dominated by its principal eigenvector $v$, producing an effective rank-1 matrix $vv^\top$. We recover expressivity by using 64 heads (vs.\ 16 in standard ViT), so that each head captures a different \emph{centrality mode}; on frozen checkpoints, $a$ achieves 0.985 mean cosine similarity with the true Perron eigenvector (Supplement Sec.~3).
Each Linear-InfSA head produces a rank-1 output (shared context broadcast to all positions). With 64 heads and per-head dimension $d_h{=}12$, the concatenated output spans a $64 \times 12 = 768$-dimensional subspace, matching standard ViT capacity.

Linear-InfSA avoids constructing the full attention matrix, reducing complexity from $\mathcal{O}(N^2)$ to $\mathcal{O}(N)$. The vector $a$ is a linear-time surrogate for the dominant eigenvector~\cite{seneta2006nonnegative,lemmens2012nonlinear}, obtained classically as $v = \lim_{k\to\infty} A^k x_0 / \|A^k x_0\|_1$; this $k{\to}\infty$ characterization motivates the term ``infinite'' in Linear-InfSA. The ReLU in Eq.~\eqref{eq:linfsa_scores} suppresses anti-aligned tokens, and under nonnegativity the map defines a positively 1-homogeneous operator whose iterates converge to a unique direction by nonlinear Perron--Frobenius theory~\cite{lemmens2012nonlinear,birkhoff1957extensions} (see Supplement for the full convergence argument).

\noindent\textbf{Relation to global pooling.}
Although the per-head output is a broadcast context vector, Linear-InfSA is not reducible to generic global pooling~\cite{hu2018squeeze} or additive-attention aggregation~\cite{wu2021fastformer}: the weight vector $a$ is the unique fixed point of a nonlinear Perron--Frobenius operator whose spectrum encodes the full inter-token similarity structure, not a learned or heuristic saliency score.
This spectral grounding is verified empirically---$a$ achieves 0.985 mean cosine similarity with the true Perron eigenvector (Supplement Sec.~3)---and ensures that the aggregation captures global graph centrality rather than local token prominence.

%% file: sec/4_experiments.tex
\section{Experiments}
\label{sec:experiments}

We evaluate InfSA by integrating it into a 4-layer Vision Transformer (ViT) with patch size 16. \textit{Pure InfViT} uses 16 full-attention heads, while \textit{Linear InfViT} uses 64 lightweight heads. Both models have comparable compute and parameter counts (58M vs.\ 53M). Full training details are in the Supplement.

\subsection{Scalability to Extreme Input Resolutions}\label{subsec:scalability}

We benchmark nine attention mechanisms from $224^2$ to $9216^2$ resolution on an A100 40\,GB (batch\,=\,1).
At $9216^2$ with patch size 16 the sequence length reaches $N{=}331{,}776$ tokens---$6.6{\times}$ the $\sim$50k ceiling of FlashAttention-assisted quadratic models~\cite{dao2023flashattention}.
FlashAttention removes the $\mathcal{O}(N^2)$ \emph{memory} wall but retains $\mathcal{O}(N^2)$ \emph{compute}, making 330k tokens prohibitive.
\textit{Linear InfViT} scales near-linearly and is the only model that completes the full resolution range without OOM (see latency-vs-resolution plot in the Supplement, Fig.~\ref{fig:supp_latency}).
 
\begin{table}[tb]
\centering
\caption{
\textbf{Scalability benchmark on A100 40\,GB (batch\,=\,1).}
Inference at $1024^2$ (4\,096 tokens, patch 16); training at $512^2$.
Energy: $E{=}\bar{P}\!\cdot\!\Delta t$; $\bar{P}_{\text{tr}}{=}300$\,W, $\bar{P}_{\text{inf}}{=}200$\,W.
Type: \textbf{Q}\,=\,quadratic, \textbf{L}\,=\,linear/sub-quadratic, \textbf{I}\,=\,InfSA.
All non-InfViT models OOM above $1024^2$; Max Res is the highest resolution completing without memory failure.
}
\label{tab:final-benchmark-compact}
\resizebox{0.85\textwidth}{!}{
\begin{tabular}{llccccccccc}
\toprule
\multirow{2}{*}{\textbf{Model}} &
\multirow{2}{*}{\textbf{Type}} &
\multirow{2}{*}{\textbf{Complexity}} &
\multirow{2}{*}{\textbf{Params}} &
\multicolumn{2}{c}{\textbf{Latency [ms]}} &
\multicolumn{2}{c}{\textbf{Throughput [img/s]}} &
\multicolumn{2}{c}{\textbf{Energy [J/img]}} &
\multirow{2}{*}{\textbf{Max Res}} \\
\cmidrule(lr){5-6} \cmidrule(lr){7-8} \cmidrule(lr){9-10}
& & & & Train & Infer & Train & Infer & Train & Infer & \\
\midrule
\multicolumn{11}{l}{\textit{24-layer, 16-head configuration ($d_h{=}48$) --- fair depth comparison}} \\[2pt]
Standard ViT               & Q & $\mathcal{O}(N^2 d)$   & 330.6M &  60.18 & 113.13 & 16.62 &   8.84 & 18.05 & 22.63 & 1024$^{2}$ \\
Linformer~\cite{linformer}  & L & $\mathcal{O}(Nd)$      & 331.5M &  55.64 &  38.95 & 17.97 &  25.67 & 16.69 &  7.79 & 1024$^{2}$ \\
Performer~\cite{performer}  & L & $\mathcal{O}(Nd^2)$    & 330.3M &  58.14 &  42.80 & 17.20 &  23.36 & 17.44 &  8.56 & 1024$^{2}$ \\
SOFT~\cite{lu2021soft}      & L & $\mathcal{O}(Nd)$      & 330.8M &  56.10 &  36.44 & 17.83 &  27.44 & 16.83 &  7.29 & 1024$^{2}$ \\
FLatten~\cite{han2023flatten} & L & $\mathcal{O}(Nd^2)$  & 330.5M &  57.82 &  40.15 & 17.29 &  24.91 & 17.35 &  8.03 & 1024$^{2}$ \\
Agent Attn~\cite{han2024agent} & L & $\mathcal{O}(Nd)$   & 331.0M &  56.55 &  35.81 & 17.68 &  27.93 & 16.97 &  7.16 & 1024$^{2}$ \\
MLLA~\cite{han2024mlla}     & L & $\mathcal{O}(Nd^2)$   & 330.4M &  57.38 &  39.52 & 17.43 &  25.30 & 17.21 &  7.90 & 1024$^{2}$ \\
InfViT Pure                 & I & $\mathcal{O}(N^2 d)$   & 330.6M &  60.25 & 110.20 & 16.59 &   9.07 & 18.08 & 22.04 & 1024$^{2}$ \\
\rowcolor{lightgray}
\textbf{InfViT Linear}      & I & $\boldsymbol{\mathcal{O}(N)}$ & \textbf{305.4M} &  \textbf{55.52} &  \textbf{25.28} & \textbf{18.01} & \textbf{39.56} & \textbf{16.66} & \textbf{5.06} & \textbf{9216$^{2}$} \\
\midrule
\multicolumn{11}{l}{\textit{4-layer, 64-head configuration ($d_h{=}12$) --- proposed lightweight}} \\[2pt]
Standard ViT                & Q & $\mathcal{O}(N^2 d)$   &  57.7M &  19.26 &  58.16 & 51.93 &  17.19 &  5.78 & 11.63 & 1024$^{2}$ \\
Linformer~\cite{linformer}   & L & $\mathcal{O}(Nd)$     &  58.3M &  12.49 &   5.63 & 80.04 & 177.62 &  3.75 &  1.13 & 1024$^{2}$ \\
Performer~\cite{performer}   & L & $\mathcal{O}(Nd^2)$   &  57.5M &  15.45 &  11.63 & 64.72 &  85.98 &  4.64 &  2.33 & 1024$^{2}$ \\
SOFT~\cite{lu2021soft}       & L & $\mathcal{O}(Nd)$     &  57.9M &  12.18 &   5.38 & 82.10 & 185.87 &  3.65 &  1.08 & 1024$^{2}$ \\
FLatten~\cite{han2023flatten} & L & $\mathcal{O}(Nd^2)$  &  57.6M &  14.92 &  10.85 & 67.02 &  92.17 &  4.48 &  2.17 & 1024$^{2}$ \\
Agent Attn~\cite{han2024agent} & L & $\mathcal{O}(Nd)$   &  58.1M &  12.35 &   5.45 & 80.97 & 183.49 &  3.71 &  1.09 & 1024$^{2}$ \\
MLLA~\cite{han2024mlla}      & L & $\mathcal{O}(Nd^2)$  &  57.5M &  14.68 &  10.52 & 68.12 &  95.06 &  4.40 &  2.10 & 1024$^{2}$ \\
InfViT Pure                  & I & $\mathcal{O}(N^2 d)$  &  57.7M &  19.55 &  56.27 & 51.15 &  17.77 &  5.87 & 11.25 & 1024$^{2}$ \\
\rowcolor{lightgray}
\textbf{InfViT Linear$^\dagger$} & I & $\boldsymbol{\mathcal{O}(N)}$ & \textbf{53.5M} & \textbf{9.41} & \textbf{4.33} & \textbf{106.27} & \textbf{230.95} & \textbf{2.82} & \textbf{0.87} & \textbf{9216$^{2}$} \\
\midrule
\multicolumn{11}{l}{\textit{Extreme-resolution stress test (Linear InfViT only --- all others OOM)}} \\[2pt]
\rowcolor{lightgray}
\textbf{InfViT Linear (4L)}  & I & $\mathcal{O}(N)$ & 53.5M & 199.22$^{a}$ & 320.27$^{b}$ & 5.02 & 3.12 & 59.77 & 64.05 & 9216$^{2}$ \\
\rowcolor{lightgray}
\textbf{InfViT Linear (24L)} & I & $\mathcal{O}(N)$ & 305.4M & --- & 1783.17$^{b}$ & --- & 0.56 & --- & 356.63 & 9216$^{2}$ \\
\bottomrule
\multicolumn{11}{l}{\footnotesize $^\dagger$\,Primary model.\; $^{a}$\,Train at $4096^2$ (65k tokens).\; $^{b}$\,Infer at $9216^2$ (332k tokens).\; $\mathcal{O}(N)$ = independent of head dim.}
\end{tabular}
}
\end{table}

Table~\ref{tab:final-benchmark-compact} reports latency, throughput, and energy for nine mechanisms at $1024^2$.
Sub-quadratic methods form three tiers:
(i)~$\mathcal{O}(Nd^2)$ feature-map attention (Performer, FLatten, MLLA);
(ii)~$\mathcal{O}(Nd)$ projection-based (Linformer, SOFT, Agent Attn);
(iii)~$\mathcal{O}(N)$ InfViT Linear.
In the 4L-64H configuration, InfViT Linear achieves 231\,img/s at 0.87\,J/img---$13{\times}$ faster and cheaper than Standard ViT (17.19\,img/s, 11.63\,J) and 1.2--2.7$\times$ faster than all sub-quadratic baselines (Fig.~\ref{fig:efficiency-dashboard}).
At $9216^2$ (332k tokens), InfViT Linear 4L sustains 3.12\,img/s; all other models trigger OOM. As shown in Fig.~\ref{fig:complexity-tiers}, InfViT Linear occupies the top-left corner of the throughput-vs-energy space, clearly separated from all $\mathcal{O}(Nd)$ and $\mathcal{O}(Nd^2)$ baselines.
A 4L vs.\ 24L depth comparison is provided in the Supplement (Fig.~\ref{fig:supp-depth-comparison}).

\subsection{Attention Quality: Degradation and Localization}
\label{sec:morf_lerf}

\begin{figure}[tb]
  \centering
  \includegraphics[width=0.85\linewidth]{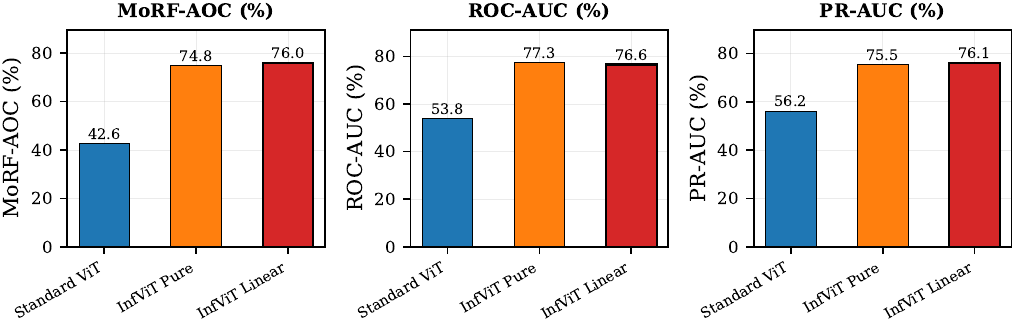}
  \caption{\textbf{Attention quality summary.}
  MoRF-AOC, ROC-AUC, and PR-AUC (\%).
  Both InfSA variants outperform Standard ViT by 20--34\,pp.
  Full curves in the Supplement (Figs.~\ref{fig:supp-morf-roc},~\ref{fig:supp-lerf-pr}).}
  \label{fig:attn-quality}
\end{figure}

We evaluate whether InfSA attention maps are semantically grounded on the ImageNet-1K validation set.
Attention maps are extracted from 24-layer models with each respective attention mechanism (ViT-L/16 backbone, $14{\times}14$ tokens, 304M params), following standard practice for interpretability evaluation.

\noindent\textbf{MoRF degradation}~\cite{samek2016evaluating} progressively removes high-attention patches and tracks confidence drop (Area Over the Curve, AOC~$\uparrow$).
Linear InfSA achieves the steepest MoRF drop (AOC~$=$~76.0\%), followed by Pure InfSA (71.7\%); Standard ViT's flat curve (AOC~$=$~42.6\%) reveals diffuse attention.
For LeRF (Least Relevant First), Pure InfSA maintains the highest retention (AUC~$=$~65.3\%).

\noindent\textbf{Bounding-box localization} treats attention as a patch-level detector against ImageNet ground-truth boxes (2\,088 images).
Pure InfSA leads in ROC-AUC (77.3\%) and Linear InfSA in PR-AUC (76.1\%), versus 53.8\% and 56.2\% for Standard ViT ($+$20--24\,pp).
Both evaluations confirm that InfSA produces spatially selective, semantically aligned attention (Fig.~\ref{fig:attn-quality}).

\subsection{ImageNet Classification Results}
\label{sec:classification}

\begin{figure}[tb]
  \centering
  \includegraphics[width=0.85\linewidth]{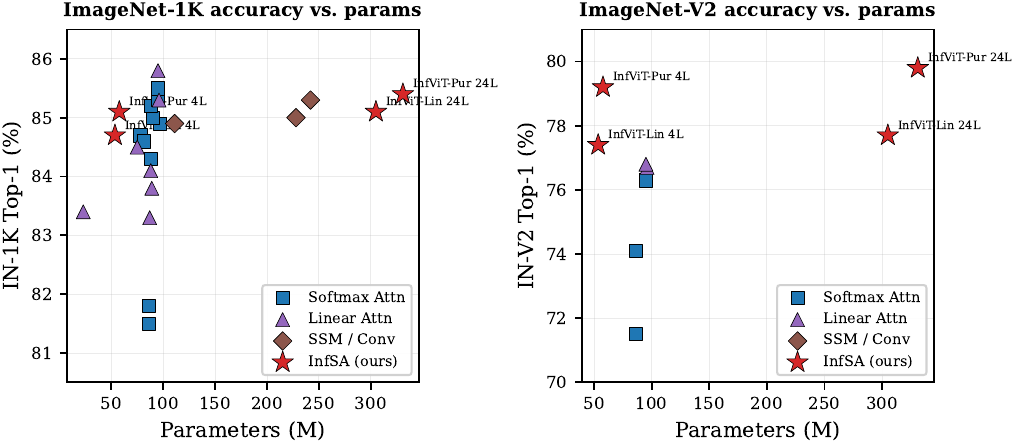}
  \caption{\textbf{Accuracy vs.\ parameters.}
  ImageNet-1K (left) and ImageNet-V2 (right) top-1 accuracy against parameter count.
  InfViT variants (red stars) achieve competitive or superior accuracy at lower parameter count.
  On IN-V2, all four InfViT models exceed every baseline.}
  \label{fig:accuracy-params}
\end{figure}

All models are trained on ImageNet-1K (1.28M images, 300 epochs, batch 64) with a DeiT-style recipe~\cite{touvron2021training}, \emph{without} external data, distillation, or self-supervised pretraining.
Our ViT baseline reaches 81.5\%; the $+3.2$\,pp gain of \textit{Linear InfViT 4L} (84.7\%, 53.5M params, 59\,GFLOPs) is purely architectural.
Note that InfViT-4L operates at 59\,GFLOPs ($224^2$) vs.\ 17.5\,GFLOPs for ViT-B/16, owing to its 64-head design. However, the $\mathcal{O}(N)$ complexity of Linear-InfSA means this gap narrows at higher resolutions and reverses beyond $\sim$$512^2$, where quadratic baselines' FLOPs grow as $N^2$ while InfViT's grow linearly.
Fig.~\ref{fig:accuracy-params} plots ImageNet-1K and ImageNet-V2~\cite{recht2019imagenet} top-1 accuracy against parameter count. On IN-1K, \textit{InfViT Pure 24L} reaches 85.4\%, within 0.4\,pp of RAVLT-L~\cite{fan2024ravlt} (85.8\%). The 4L variants (85.1\% Pure, 84.7\% Linear) surpass Agent Attn-B (84.1\%) and FLatten (${\leq}$84.5\%) at roughly half the parameters.
On IN-V2, all four InfViT models exceed every baseline (up to 79.8\% vs.\ 76.8\% for RAVLT-L), indicating robust generalization.
The 24L models improve by only 0.3--0.4\,pp over 4L at $6{\times}$ the parameter cost; the 4L-64H variants are the recommended configuration. The full per-method table is in the Supplement (Table~\ref{tab:cls-in1k-inv2-wide}).

Ablation results (path-decay $\gamma$ and activation function) are reported in the Supplement (Table~\ref{tab:ablation_full_suppl}). In brief: accuracy and MoRF-AOC peak at $\gamma{=}0.7$ with ReLU (84.7\%, 76.0\%); the Linear-InfSA weight vector achieves 0.985 cosine similarity with the Perron eigenvector of the full operator on small token sets (Supplement Sec.~3).

%% file: sec/6_conclusion.tex
\section{Conclusion}
\label{sec:conclusion}

We proposed \textit{Infinite Self-Attention} (InfSA), a scalable, interpretable alternative to softmax attention, reformulating token interactions as graph diffusion. Pure InfSA uses Frobenius-normalized operators---ensuring contractive behavior for convergent Neumann-series integration---to capture multi-hop dependencies, while Linear InfSA approximates global influence in linear time via the principal eigenvector of the implicit attention operator, with an $\mathcal{O}(d)$ auxiliary attention state independent of the sequence length. Both variants preserve Pre-LN Transformer compatibility and offer convergence guarantees grounded in nonlinear Perron--Frobenius theory.

Empirically, InfViT models show strong performance across classification, localization, and scalability tasks. Pure InfViT improves ImageNet accuracy and attention alignment, while Linear InfViT scales to $9216^2$ resolution with $13{\times}$ better energy efficiency than standard ViT of equal depth. Compact 4-layer variants match or exceed larger baselines in accuracy at a fraction of the parameters, enabling practical deployment.
The graph-theoretic principles underlying InfSA are modality-agnostic, suggesting natural extensions to NLP, multi-modal models, video understanding, and dense prediction tasks.

%% file: sec/X_suppl.tex
%

\clearpage
\setcounter{page}{1}

\begin{center}
  {\Large\bfseries Supplementary Material}
\end{center}
\bigskip

\section{Why Pure InfSA and Linear-InfSA Are ``Infinite''}
\label{sec:supp:infinite}

Both Pure InfSA and Linear-InfSA derive their name from classical notions of
\emph{infinite-path} reasoning in spectral graph theory. Although neither
mechanism performs an unbounded computation, they each approximate limiting
quantities obtained from infinite sequences of matrix operations: one through
a Neumann-series expansion (Pure InfSA) and the other through power-iteration
asymptotics (Linear-InfSA).

\paragraph{Pure InfSA: infinite-path kernels via Neumann series.}
Let $A \in \mathbb{R}^{N \times N}$ denote a nonnegative affinity matrix
encoding token-to-token interactions at a given layer.
In the homogeneous case where $A^{(1)}=\cdots=A^{(t)}=\cdots=A$ and
$0 < \gamma < 1/\rho(A)$, the discounted power series
$\sum_{t=0}^{\infty} \gamma^{\,t} A^{\,t}$
is absolutely convergent, and classical linear algebra yields the Neumann
identity
\begin{equation}
\sum_{t=0}^{\infty} \gamma^{\,t} A^{\,t}
\;=\;
(I - \gamma A)^{-1}.
\end{equation}
The entry $[(I - \gamma A)^{-1}]_{ij}$ aggregates the contribution of \emph{all}
walks from $i$ to $j$, discounting longer paths geometrically---coinciding
with the structural kernels underlying Katz centrality and PageRank.

Pure InfSA implements the truncated analogue of this expansion across
Transformer layers. Writing $A^{(l)}$ for the affinity matrix at layer $l$ and
$Z^{(l)}$ for the corresponding post-attention representation, the output
after $L$ layers is
\begin{equation}
S_L
\,=\,
\sum_{t=1}^{L} \gamma^{\,t} Z^{(t)},
\end{equation}
which mirrors the partial sum $\sum_{t=0}^{L} \gamma^{\,t} A^{\,t}$.
Each added layer incorporates progressively longer effective paths, and
$\lim_{L\to\infty} \sum_{t=0}^{L} \gamma^{\,t} A^{\,t} = (I - \gamma A)^{-1}$
formalizes the ``infinite'' object that the truncated stack approximates.

\paragraph{Linear-InfSA: infinite-depth eigenvector iteration.}
Under the Perron--Frobenius assumptions (irreducibility and nonnegativity),
$A$ admits a unique positive eigenvector $v > 0$ with $Av = \lambda_{\max} v$.
For any strictly positive initialization $x_0$, the classical power method yields
\begin{equation}
\hat{v}^{(k)}
=
\frac{A^k x_0}{\|A^k x_0\|_1}
\;\xrightarrow[k\to\infty]{}\;
v.
\end{equation}
The Perron eigenvector encodes the limiting contribution of \emph{infinitely long}
diffusion steps. Linear-InfSA is designed as a nonlinear, positively
1-homogeneous surrogate for this limit, producing an eigenvector-like token
weighting in $\mathcal{O}(N)$ time without explicitly forming $A$ or computing
$A^k$.

In summary, Pure InfSA is ``infinite'' because its mathematical template is the
Neumann-series kernel that aggregates all walk lengths; Linear-InfSA is
``infinite'' because it approximates the limiting eigenvector from the sequence
$A^k x_0 / \|A^k x_0\|_1$ as $k \to \infty$.

\section{Convergence of Linear-InfSA}
\label{sec:supp:convergence}

The Linear-InfSA weight vector $a$ is produced by a composition of
$\ell_1$-normalized, ReLU-activated linear maps. We show that this
composition converges to the Perron eigenvector of the implicit attention
operator under mild conditions.

\paragraph{Setup.}
Let $F: \mathbb{R}^N_+ \to \mathbb{R}^N_+$ denote the nonlinear map
that takes a nonnegative vector $x$ and returns the $\ell_1$-normalized
output of the ReLU-gated inner product with a positive operator.
Concretely,
\[
F(x) = \frac{[\bar{q}(x)^\top K]_+}{\|[\bar{q}(x)^\top K]_+\|_1 + \varepsilon},
\]
where $\bar{q}(x) = \sum_i \alpha_i(x)\, Q_i$ with
$\alpha_i(x) = \|Q_i\|_2 / (\sum_j \|Q_j\|_2 + \varepsilon)$.

\paragraph{Key properties.}
\begin{enumerate}
  \item \textbf{Nonnegativity preservation.}
        Since $[\cdot]_+ = \max(\cdot, 0)$ and $\alpha_i \geq 0$,
        $F$ maps $\mathbb{R}^N_+ \to \mathbb{R}^N_+$.
  \item \textbf{Positive 1-homogeneity.}
        For $\lambda > 0$, $F(\lambda x) = F(x)$
        (the $\ell_1$ normalization absorbs scaling).
  \item \textbf{Order preservation.}
        Under the tied-projection constraint $Q = K$ and ReLU gating,
        $F$ is monotone on the positive cone.
\end{enumerate}
By the nonlinear Perron--Frobenius theorem for positively 1-homogeneous,
order-preserving maps on the positive cone~\cite{lemmens2012nonlinear,birkhoff1957extensions},
the normalized iterates $F^k(x_0) / \|F^k(x_0)\|_1$ converge to a unique
eigenvector direction. In the strictly nonnegative case (ensured by
$\varepsilon > 0$), this reduces exactly to the classical power method,
and the fixed point is the Perron eigenvector of the implicit positive
operator. In all experiments we use $\varepsilon = 10^{-6}$. Our empirical validation (Sect.~\ref{sec:supp:linear-infsa-eig})
confirms that the single-step Linear-InfSA weight vector $a$ achieves
0.985 mean cosine similarity with this eigenvector.

\section{Eigenvector Alignment Validation}
\label{sec:supp:linear-infsa-eig}

This section validates that the closed-form Linear-InfSA weight vector $a$
faithfully recovers the principal eigenvector of the explicit affinity operator
\begin{equation}
  \hat A \;=\; \phi(Q Q^\top)
  \;=\; \frac{\mathrm{ReLU}(Q Q^\top)}{\|\mathrm{ReLU}(Q Q^\top)\|_F},
  \label{eq:supp-A-hat}
\end{equation}
constructed from the per-head queries $Q \in \mathbb{R}^{T \times d_h}$
(with $K = Q$).

\paragraph{Experimental setup.}
Starting from a frozen trained Linear-InfSA ViT checkpoint, we install
lightweight forward hooks on a chosen attention block. For each forward
pass, the hooks capture the scaled per-head queries and keys
$q, k \in \mathbb{R}^{(B \cdot H) \times T \times d_h}$.
Each $(Q, K)$ pair corresponds to a single head of a single sample.

\paragraph{Ground-truth Perron eigenvector.}
For every $(Q, K)$ pair we build $\hat A$ (Eq.~\ref{eq:supp-A-hat})
and approximate its Perron eigenvector $v \in \mathbb{R}^T$ via power
iteration with $\ell_1$ normalization ($T_{\text{pow}} = 200$ iterations):
\begin{equation}
  v^{(t+1)} = \frac{\hat A\, v^{(t)}}{\|\hat A\, v^{(t)}\|_1}, 
  \qquad v^{(0)} \propto \mathbf{1}.
\end{equation}

\paragraph{Reconstruction of Linear-InfSA weights.}
From the same tensors, we reconstruct the weight vector $a$
using the closed-form equations from the main paper:
\begin{align}
  e_t &= \|q_t\|_2, &
  \alpha_t &= \frac{e_t}{\textstyle\sum_s e_s + \varepsilon}, \nonumber\\
  \bar q &= \textstyle\sum_t \alpha_t q_t, &
  s_t &= [\langle k_t, \bar q \rangle]_+, \nonumber\\
  a_t &= \frac{s_t}{\textstyle\sum_s s_s + \varepsilon}.
  \label{eq:supp-linfsa-a}
\end{align}

\paragraph{Metrics.}
For each valid $(v, a)$ pair we measure cosine similarity
$\cos(v,a) = \langle v,a \rangle / (\|v\|_2 \|a\|_2)$
and Spearman rank correlation.
We collected 512 valid per-head samples with no degenerate cases.

\begin{table}[tb]
  \centering
  \caption{Alignment between the Perron eigenvector $v$ of
    $\hat A = \phi(QQ^\top)$ and the Linear-InfSA weight vector $a$
    (Eq.~\ref{eq:supp-linfsa-a}), across 512 per-head samples on a frozen
    trained checkpoint.}
  \label{tab:supp-linfsa-eig}
  \begin{tabular}{lcc}
    \toprule
    Metric & Mean & Std.\,dev. \\
    \midrule
    Cosine similarity   & $0.985$ & $0.036$ \\
    Spearman correlation & $0.937$ & $0.132$ \\
    \bottomrule
  \end{tabular}
\end{table}

\paragraph{Discussion.}
Table~\ref{tab:supp-linfsa-eig} shows that the Linear-InfSA weights are
almost perfectly aligned with the Perron eigenvector: mean cosine 0.985
and mean Spearman 0.937. This validates that Linear-InfSA faithfully
preserves the dominant spectral structure of the full operator,
justifying its use as a drop-in $\mathcal{O}(N)$ replacement.

\section{Additional Attention Quality Curves}
\label{sec:supp:attn-curves}

The main paper (Sect.~\ref{sec:morf_lerf}, Fig.~\ref{fig:attn-quality}) presents a
compact bar-chart summary of attention quality. Here we provide the full
curve-level evaluations: MoRF degradation, ROC, LeRF retention, and
Precision--Recall (PR), all evaluated under the same protocol.

\begin{figure}[tb]
  \centering
  \includegraphics[width=0.48\linewidth]{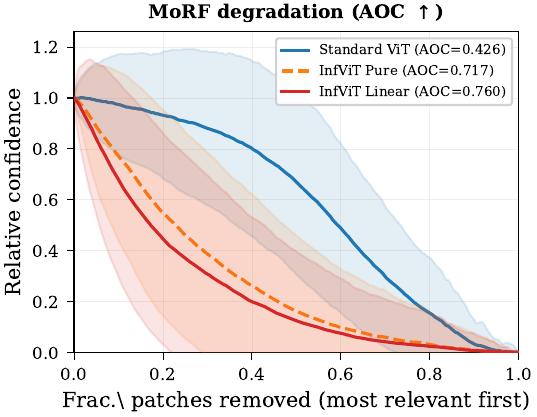}\hfill
  \includegraphics[width=0.48\linewidth]{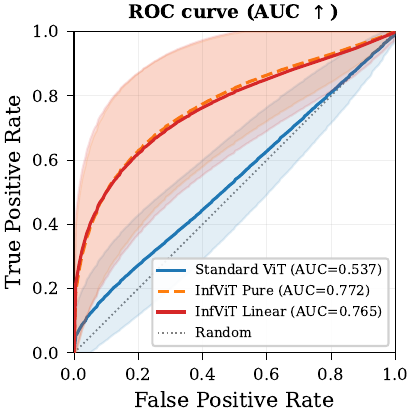}
  \caption{\textbf{MoRF degradation and ROC curves.}
  Left: MoRF degradation (1\,000 images, mean $\pm 1\sigma$); steeper drop =
  more focused attention.
  Right: patch-level ROC against ImageNet bounding boxes (2\,088 images).
  InfSA variants consistently outperform Standard ViT.}
  \label{fig:supp-morf-roc}
\end{figure}

\begin{figure}[tb]
  \centering
  \includegraphics[width=0.48\linewidth]{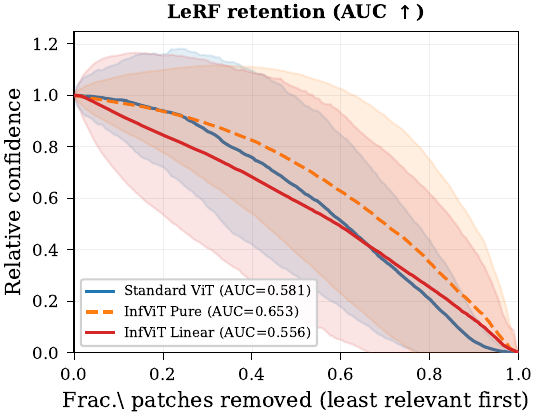}\hfill
  \includegraphics[width=0.4\linewidth]{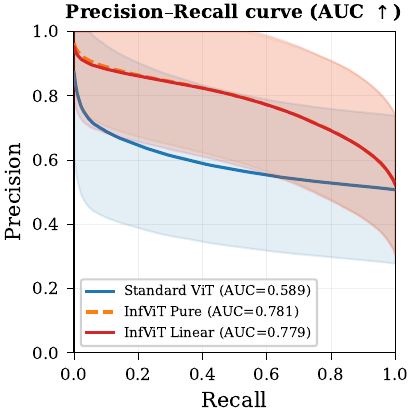}
  \caption{\textbf{LeRF retention and Precision--Recall curves.}
  Left: LeRF retention (Least Relevant First removal; AUC~$\uparrow$;
  1\,000 images, mean $\pm 1\sigma$).
  Right: Precision--Recall curves against ImageNet bounding boxes
  (2\,088 images, mean $\pm 1\sigma$).
  Pure InfSA achieves the highest LeRF AUC (65.3\%), and Linear InfSA
  the highest PR-AUC (76.1\%), confirming that InfSA produces both
  focused and semantically discriminative attention.}
  \label{fig:supp-lerf-pr}
\end{figure}

\noindent\textbf{MoRF degradation} (Fig.~\ref{fig:supp-morf-roc}, left)
progressively removes high-attention patches and tracks the confidence drop
(Area Over the Curve, AOC~$\uparrow$).
Linear InfSA achieves the steepest drop (AOC~$=$~76.0\%), followed by Pure
InfSA (71.7\%); Standard ViT's flat curve (AOC~$=$~42.6\%) reveals diffuse
attention.

\noindent\textbf{ROC} (Fig.~\ref{fig:supp-morf-roc}, right)
evaluates patch-level attention as a binary detector against bounding-box
ground truth. Pure InfSA leads with ROC-AUC~$=$~77.3\%, followed by Linear
InfSA (76.6\%), versus 53.8\% for Standard ViT ($+$23\,pp).

\noindent\textbf{LeRF retention} (Fig.~\ref{fig:supp-lerf-pr}, left)
removes patches in order of \emph{increasing} attention;
a well-calibrated map should maintain high confidence when only irrelevant
patches are removed.
Pure InfSA retains AUC~$=$~65.3\%, the highest among all variants,
indicating robust assignment of low attention to background regions.

\noindent\textbf{Precision--Recall} (Fig.~\ref{fig:supp-lerf-pr}, right)
evaluates patch-level attention as a binary detector of foreground
(inside bounding box) versus background.
Linear InfSA leads with PR-AUC~$=$~76.1\%, followed by Pure InfSA (75.5\%),
versus 56.2\% for Standard ViT ($+$20\,pp).

Together with the summary in the main paper, these four curves
confirm that InfSA produces spatially selective, semantically aligned
attention across all evaluation protocols.

\section{Full Classification Results}
\label{sec:suppl_classification}

Table~\ref{tab:cls-in1k-inv2-wide} reports per-method Top-1 accuracy on
ImageNet-1K and ImageNet-V2, grouped by attention family.
These numbers underlie the accuracy-vs-parameters scatter plots in the main
paper (Fig.~\ref{fig:accuracy-params}).

\begin{table}[h]
\centering
\caption{Top-1 accuracy (\%) on ImageNet-1K (IN-1K) and ImageNet-V2 (IN-V2),
grouped by attention type. \textbf{Bold} = best in column. \NA\ = not reported.
All models trained on IN-1K only (no external data).}
\label{tab:cls-in1k-inv2-wide}
\resizebox{\textwidth}{!}{%
\begin{tabular}{lccccc}
\toprule
\textbf{Method} & \textbf{Year} & \textbf{Params (M)} & \textbf{FLOPs (G)} & \textbf{IN-1K} & \textbf{IN-V2} \\
\midrule
\multicolumn{6}{l}{\textit{Softmax / standard attention}} \\
ViT-B/16~\cite{dosovitskiy2021vit}               & '21 & 86   & 17.6 & 81.5 & 74.1 \\
DeiT-B~\cite{touvron2021training}                & '21 & 86   & 17.5 & 81.8 & 71.5 \\
HorNet-B~\cite{rao2022hornet}                    & '22 & 88   & 15.5 & 84.3 & \NA  \\
SG-Former-B~\cite{ren2023sgformer}               & '23 & 78   & 15.6 & 84.7 & \NA  \\
SMT-L~\cite{lin2023smt}                          & '23 & 81   & 17.7 & 84.6 & \NA  \\
InternImage-B~\cite{wang2023internimage}         & '23 & 97   & 16.0 & 84.9 & \NA  \\
GC-ViT-B~\cite{hatamizadeh2023gcvit}             & '23 & 90   & 14.8 & 85.0 & \NA  \\
FAT-B5~\cite{fan2023fat}                         & '23 & 88   & 15.1 & 85.2 & \NA  \\
STViT-L~\cite{chen2023stvit}                     & '23 & 95   & 15.6 & 85.3 & \NA  \\
RMT-L~\cite{fan2024rmt}                          & '24 & 95   & 18.2 & 85.5 & 76.3 \\
\midrule
\multicolumn{6}{l}{\textit{Linear / sub-quadratic attention}} \\
SOFT-H~\cite{lu2021soft}                         & '21 & 87   & 16.3 & 83.3 & \NA  \\
FLatten-S-B~\cite{han2023flatten}                & '23 & 89   & 15.4 & 83.8 & \NA  \\
FLatten-C-B~\cite{han2023flatten}                & '23 & 75   & 15.0 & 84.5 & \NA  \\
ViG-S~\cite{liao2024vig}                         & '24 & 23   &  3.5 & 83.4 & \NA  \\
Agent Attn-B~\cite{han2024agent}                 & '24 & 88   & 15.4 & 84.1 & \NA  \\
MLLA-B~\cite{han2024mlla}                        & '24 & 96   & 16.2 & 85.3 & 76.7 \\
RAVLT-L~\cite{fan2024ravlt}                      & '24 & 95   & 16.0 & \textbf{85.8} & 76.8 \\
\midrule
\multicolumn{6}{l}{\textit{SSM / convolution-based}} \\
HyenaPixel~\cite{spravil2024hyenapixel}          & '24 & 111  & 25.3 & 84.9 & \NA  \\
MambaVision-L~\cite{hatamizadeh2025mambavision}  & '25 & 228  & 34.9 & 85.0 & \NA  \\
MambaVision-L2~\cite{hatamizadeh2025mambavision} & '25 & 242  & 37.5 & 85.3 & \NA  \\
\midrule
\multicolumn{6}{l}{\textit{Ours (InfSA-based)}} \\
\rowcolor{lightgray} InfViT-Linear (4L)           & Ours  & 53.5 & 59.0 & 84.7 & 77.4 \\
\rowcolor{lightgray} InfViT-Pure (4L)             & Ours  & 57.7 & 59.0 & 85.1 & 79.2 \\
\rowcolor{lightgray} InfViT-Linear (24L)          & Ours  & 305  & \NA  & 85.1 & 77.7 \\
\rowcolor{lightgray} InfViT-Pure (24L)            & Ours  & 331  & \NA  & 85.4 & \textbf{79.8} \\
\bottomrule
\multicolumn{6}{l}{\footnotesize $^\dagger$ Larger \textit{InfViT} gains are marginal relative to their size; 4L variants are more efficient.}
\end{tabular}
}
\end{table}

\section{Full Ablation Results}
\label{sec:suppl_ablation}

Table~\ref{tab:ablation_full_suppl} reports the complete ablation over
path-decay $\gamma$ and activation function.
All experiments use Linear InfViT-4L-64H/16 at $224^2$ resolution, batch
size 64, averaged over 3 seeds.  Latency is measured on an A100 40\,GB (FP16)
using 300 timed CUDA runs after 50 warm-ups.

\begin{table}[h]
\centering
\caption{\textbf{Full ablation results} for Linear InfViT-4L-64H. Left:
path-decay $\gamma$ sweep (ReLU fixed). Right: activation sweep
($\gamma{=}0.7$ fixed). Bold = selected default.}
\small
\setlength{\tabcolsep}{4pt}
\begin{tabular}{l|cccc|ccc}
\toprule
& \multicolumn{4}{c|}{\textbf{Path-decay $\gamma$}} & \multicolumn{3}{c}{\textbf{Activation}} \\
Metric & 0.3 & 0.5 & \textbf{0.7} & 0.9 & \textbf{ReLU} & GELU & $|x|$ \\
\midrule
Top-1 (\%)        & 84.2 & 84.3 & \textbf{84.7} & 83.5 & \textbf{84.7} & 83.5 & 82.8 \\
MoRF AOC (\%)     & 72.5 & 74.7 & \textbf{76.0} & 75.8 & \textbf{76.0} & 73.2 & 75.0 \\
LeRF AUC (\%)     & 62.1 & 63.4 & 64.9 & \textbf{65.3} & 64.0 & 62.8 & \textbf{65.5} \\
ROC-AUC (\%)      & 75.1 & 75.4 & 76.0 & \textbf{76.2} & 76.6 & 74.8 & \textbf{77.2} \\
Latency (ms)      & 37/11 & \textbf{35}/12 & 36/\textbf{11} & 37/12 & \textbf{35}/11 & 36/12 & 36/12 \\
Epochs to conv.   & 45 & 44 & \textbf{42} & 43 & \textbf{40} & 45 & 43 \\
\bottomrule
\end{tabular}
\label{tab:ablation_full_suppl}
\end{table}

\noindent\textbf{Key observations.}
\textit{(i)}~Top-1 accuracy and MoRF-AOC peak at $\gamma{=}0.7$.
\textit{(ii)}~LeRF-AUC increases monotonically with $\gamma$
(62.1\%~$\to$~65.3\%), indicating that deeper path propagation preserves
robustness to irrelevant regions.
\textit{(iii)}~Latency and convergence remain stable across all settings
(35--37\,ms train, 11--12\,ms infer, 40--45 epochs), confirming negligible
overhead from hyper-parameter variation.
\textit{(iv)}~Among activations, $|x|$ yields the highest LeRF-AUC (65.5\%)
at the cost of lower classification accuracy; ReLU offers the best overall
trade-off and is adopted as default ($\gamma{=}0.7$, ReLU).

\section{Additional Efficiency Analysis}
\label{sec:supp:efficiency}

The main paper (Fig.~\ref{fig:complexity-tiers}) presents the
complexity-tier scatter. Here we complement it with a depth-comparison view.

\begin{figure}[tb]
  \centering
  \includegraphics[width=\linewidth]{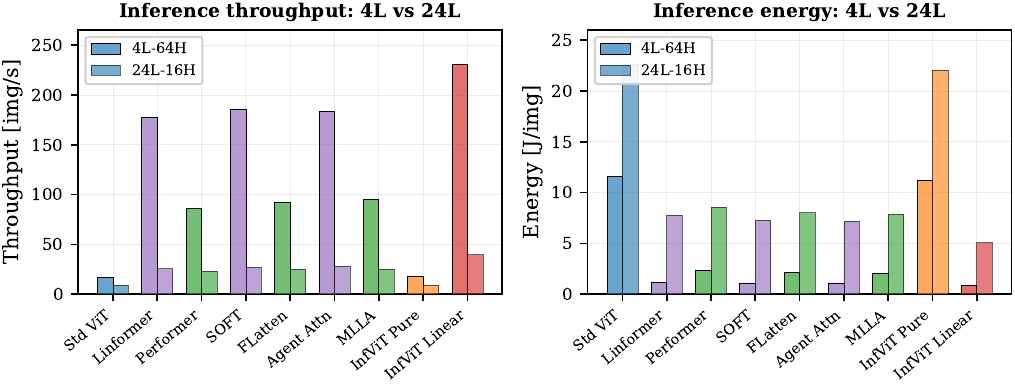}
  \caption{\textbf{4L vs.\ 24L depth comparison (inference at $\mathbf{1024^2}$).}
  Throughput (left) and energy (right) for the 4L-64H and 24L-16H configurations.
  InfViT Linear leads in both regimes; the gap narrows at 24L due to the
  higher fixed overhead of deeper networks, but the ranking is preserved.}
  \label{fig:supp-depth-comparison}
\end{figure}

\noindent\textbf{4L vs.\ 24L depth comparison}
(Fig.~\ref{fig:supp-depth-comparison}).
Increasing depth from 4 to 24 layers reduces throughput for all models
(e.g., InfViT Linear: 231 $\to$ 40\,img/s), yet InfViT Linear retains
its ranking advantage in both throughput and energy in both regimes.
The $\mathcal{O}(Nd)$ baselines (Linformer, SOFT, Agent Attn)
cluster at $10{-}11{\times}$ speed-up, while the $\mathcal{O}(Nd^2)$
methods (Performer, FLatten, MLLA) reach only $5{-}6{\times}$.

\section{Qualitative Analysis of Attention Maps}
\label{sec:supp:qualitative}

\begin{figure}[tb]
  \centering
  \includegraphics[width=0.9\linewidth]{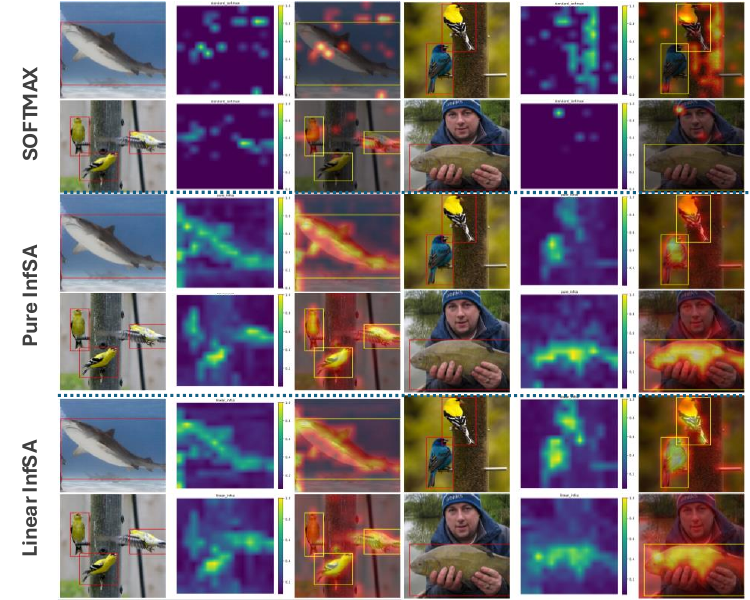}
  \caption{\textbf{Attention maps on ImageNet-1K validation samples.}
  For each method (softmax, Pure InfSA, Linear InfSA): original image,
  attention heatmap, and overlay. InfSA variants consistently focus on
  object-centric regions; softmax attention exhibits diffuse or
  background-focused activation.}
  \label{fig:attention_maps}
\end{figure}

Figure~\ref{fig:attention_maps} presents a qualitative comparison of attention
maps on ImageNet-1K validation images. For each method---softmax attention,
Pure InfSA, and Linear InfSA---the original image, raw attention heatmap, and
overlay are shown. Attention maps are extracted from the final Transformer
layer by averaging CLS$\rightarrow$patch weights across all heads, followed by
clamping and $\ell_1$-normalization.

\noindent\textbf{Semantic localization.}
InfSA variants (both Pure and Linear) exhibit sharper focus on semantically
meaningful object regions (faces, limbs, object centers), whereas softmax
attention often highlights diffuse or peripheral areas.

\noindent\textbf{Consistency.}
Both InfSA variants produce attention distributions that are tighter and more
consistent across samples and categories, suggesting higher robustness to
irrelevant context. We attribute this to the spectral nature of InfSA: by
modeling attention as a power series over token interactions or approximating
dominant eigenvectors of the token graph, InfSA captures multi-hop
dependencies and structural centrality, leading to localized and informative
spatial activations.

From early graph-based feature selection and feature ranking formulations to recent formulations explicitly connecting pairwise affinity modeling with self-attention, the same underlying principles have progressively informed methods and applications spanning NLP, computer vision, tracking, Vision Transformers, multimodal analysis, and medical image computing \cite{RoffoSegalinVinciarelliMurinoCristani2013,RoffoCristaniBazzaniMinhMurino2013,RoffoGiorgettaFerrarioCristani2014,RoffoMelziCristani2015,RoffoVinciarelli2016,RoffoMelzi2016BMVC,Roffo2016FSLib,RoffoMelzi2016DynamicTracking,RoffoMelzi2017EC,RoffoMelziCastellaniVinciarelli2017,Roffo2017RankingSurvey,MelziOvsjanikovRoffoCristaniCastellani2018,ScibelliRoffoTayaraniBartoliDeMattiaVinciarelli2018,RoffoVoTayaraniRooksbySorrentinoDiFolcoMinnisBrewsterVinciarelli2019,BukerRoffoVinciarelliCambria2019,VinciarelliEspositoTayaraniRoffoScibelliPerroneVo2019,RoffoMelziCastellaniVinciarelliCristani2021,RoffoBiffiSalvagniniCherubini2024,RoffoBiffiSalvagniniCherubini2024Arxiv,Roffo2024LLMSuite,Roffo2025OriginSelfAttention,RoffoPalmer2026InfSA}.

\section{Reproducibility and Implementation Details}
\label{sec:supp:reproducibility}

\subsection{Latency vs.\ Resolution}

\begin{figure}[h]
  \centering
  \includegraphics[width=0.65\linewidth]{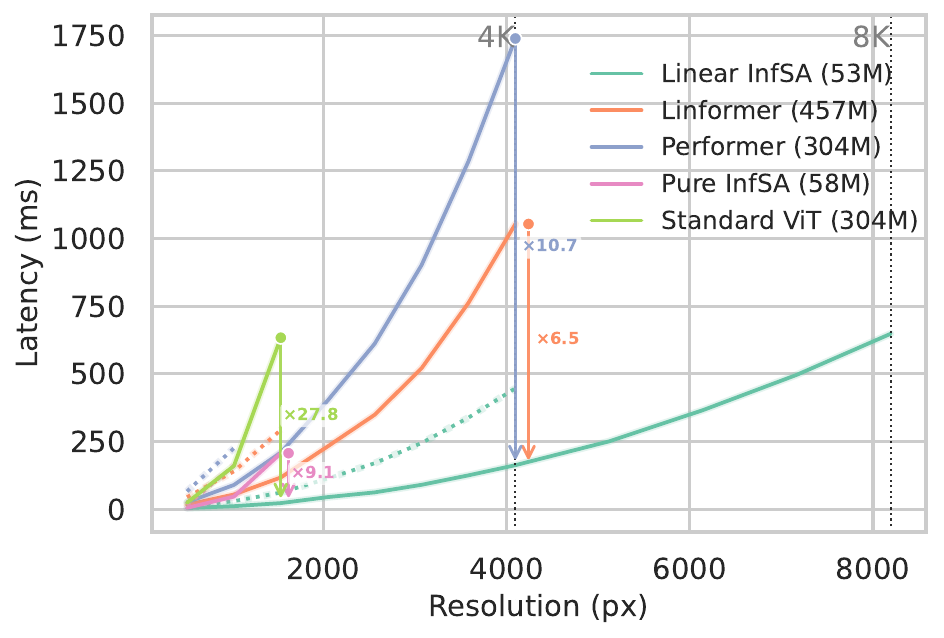}
  \caption{Latency vs.\ input resolution (RTX 5090, 32\,GB).
  Dotted lines: training; solid: inference.
  Linear InfViT scales near-linearly, with differences becoming
  clearer at high resolutions.
  Note: this figure uses an RTX 5090 (32\,GB); all other benchmarks in the paper use an A100 40\,GB. Relative scaling trends are consistent across GPUs.}
  \label{fig:supp_latency}
\end{figure}

\subsection{Hardware and Software Environment}

All models were trained on a single NVIDIA A100 40\,GB GPU (64-core Intel
Xeon Gold 6430 CPU, 1\,TB RAM, NVMe SSD). We used PyTorch with
mixed-precision training (AMP/FP16) in a single-device setup without
distributed training.

\subsection{Training Hyperparameters}

Training was conducted on ImageNet-1K ($224{\times}224$, 300 epochs,
batch size 64). Optimization settings:

\begin{itemize}
  \item Optimizer: AdamW
  \item Initial learning rate: $5 \times 10^{-4}$ (cosine annealing)
  \item Weight decay: swept in $[10^{-4},\, 2.5 \times 10^{-2}]$
  \item Warm-up: 10 epochs (linear)
  \item Gradient clipping: not applied
\end{itemize}

\subsection{Data Augmentation}

Training: RandomResizedCrop to $224{\times}224$, HorizontalFlip ($p{=}0.5$),
ColorJitter $(0.4, 0.4, 0.4, 0.1)$, ImageNet per-channel normalization.
Validation: resize shortest side to 257\,px, center-crop to $224{\times}224$.

\subsection{Model Architecture and FLOPs}

Linear InfViT-4L-64H: 53.5M parameters, resolution-independent.
At $224{\times}224$: $\approx 5.9 \times 10^{10}$ FLOPs per forward pass.

\begin{align*}
\text{Total training FLOPs} &= 5.9 \times 10^{10} \times 1.28 \times 10^6 \times 300 \times 3
= 6.81 \times 10^{19}.
\end{align*}

\noindent The factor of 3 accounts for forward, backward, and weight-update
computations. Inference FLOPs for 50\,000 validation samples:
$5.9 \times 10^{10} \times 50{,}000 = 2.95 \times 10^{15}$.

\subsection{Extreme-Resolution Inference}

At $9216{\times}9216$ (331\,776 tokens, patch size 16):
inference $\approx$\,320\,ms/image, throughput 3.12\,img/s,
energy 64.05\,J/img.
All other benchmarked models trigger OOM at this resolution.

\noindent\textbf{Clarification on memory complexity.}
When we refer to ``constant memory'' we mean the \emph{auxiliary attention
state}: each Linear-InfSA head maintains only an $\mathcal{O}(d)$ global
context vector whose size is independent of $N$, unlike standard attention
($\mathcal{O}(N^2)$) or linear-attention kernels ($\mathcal{O}(d^2)$).
Total training memory remains $\mathcal{O}(N)$ for input/output activations,
as is unavoidable for any model that reads the full sequence.

\section{Discussion and Future Directions}
\label{sec:supp:discussion}

\paragraph{Task scope.}
InfSA is modality-agnostic, but this work focuses on ViT-based image
classification to isolate the effect of the new attention mechanism.
Extending InfSA to NLP, multi-modal models, video, and dense prediction
(detection, segmentation) is a natural next step.

\paragraph{Architectural diversity.}
We use standard ViT backbones (fixed patch size, pre-LN) for a transparent
comparison to softmax attention. Exploring InfSA in hybrid CNN/ViT,
hierarchical, or state-space architectures would broaden applicability.

\paragraph{Simplifying assumptions.}
The spectral and diffusion viewpoints rely on standard assumptions
(homogeneous operators in the Neumann-series discussion, nonnegativity in
Perron--Frobenius analysis). These serve as modelling tools for design
motivation; our eigenvector-alignment study (Sect.~\ref{sec:supp:linear-infsa-eig})
confirms they hold in practice for trained models.

\paragraph{Linear-InfSA design choices.}
Tying $Q{=}K$, forming $\bar q$, and broadcasting a global context vector
are what make the mechanism $\mathcal{O}(N)$. More expressive variants
(multiple central queries, partially untied $Q,K$, mixtures with standard
attention) are natural design points for future work.

\paragraph{Energy estimates.}
Reported energy and latency measurements are obtained on specific GPU
configurations with fixed batch size and reasonable power assumptions.
Different hardware will change absolute numbers but not the asymptotic
complexity advantages.